\documentclass{article}

\usepackage[preprint]{neurips_2023}

\usepackage{neurips_2023}
\usepackage[utf8]{inputenc} 
\usepackage[T1]{fontenc}    
\usepackage{hyperref}       
\usepackage{url}            
\usepackage{amsfonts}       
\usepackage{nicefrac}       
\usepackage{microtype}      
\usepackage{xcolor}         
\usepackage{amsmath}
\usepackage{amssymb}
\usepackage{algorithm}
\usepackage{algorithmic}
\usepackage{mathrsfs}
\usepackage{ulem} 
\usepackage{cancel}
\usepackage{caption}
\usepackage{subcaption}
\usepackage{multirow}
\usepackage{color}
\usepackage{tabularx}
\usepackage{arydshln}
 \usepackage{graphicx}
\usepackage{enumitem}
\usepackage{makecell}
\usepackage{subcaption}
\usepackage[figuresright]{rotating}
\useunder{\uline}{\ul}{}

\usepackage{float}
\newfloat{figtab}{htb}{fgtb}
\makeatletter
  \newcommand\figcaption{\def\@captype{figure}\caption}
  \newcommand\tabcaption{\def\@captype{table}\caption}
\makeatother

\usepackage{multirow}

\def\T{{ \mathrm{\scriptscriptstyle T} }}
\renewcommand{\top}{\T}

\def\$#1\${\begin{align*}#1\end{align*}}

\usepackage{minitoc}

\title{Directional diffusion models for graph representation learning}

\author{%
  Run Yang\\
  Department of AI PlatForm \\
  Baidu\\
  \texttt{yangrun@baidu.com} \\
  \And
  Yuling Yang \\
  Department of Statistics and Management \\
  Shanghai University of Finance and Economics \\
  \texttt{sibyllayang@163.sufe.edu.cn} \\
    \And
  Fan Zhou \\
  Department of Statistics and Management \\
  Shanghai University of Finance and Economics \\
  \texttt{zhoufan@mail.shufe.edu.cn} \\
    \And
  Qiang Sun \\
  Department of Statistical Sciences\\
  University of Toronto \\
  \texttt{qiang.sun@utoronto.ca} \\
}

\begin{document}

\maketitle

\begin{abstract}

In recent years, diffusion models have achieved  remarkable success in various domains of artificial intelligence, such as image synthesis, super-resolution, and 3D molecule generation. However, the application of diffusion models in graph learning has received relatively little attention. In this paper, we address this gap by investigating the use of diffusion models for unsupervised graph representation learning.  We begin by identifying the anisotropic structures of graphs and a crucial limitation of the vanilla forward diffusion process in learning anisotropic structures. This process relies on continuously adding an isotropic Gaussian noise to the data, which may convert the anisotropic signals to noise too quickly. This rapid conversion hampers the training of denoising neural networks and impedes the acquisition of semantically meaningful representations in the reverse process. To address this challenge, we propose a new class of models called {\it directional diffusion models}. These models incorporate data-dependent, anisotropic, and  directional noises in the forward diffusion process. To assess the efficacy of our proposed models, we conduct extensive experiments on 12 publicly available datasets, focusing on two distinct graph representation learning tasks. The experimental results  demonstrate the superiority of our models over state-of-the-art baselines, indicating their effectiveness in capturing meaningful graph representations. Our studies  not only provide valuable insights into the forward process of diffusion models but also highlight the wide-ranging potential of these models for various graph-related tasks.
\end{abstract}

\vspace{-30pt}
\doparttoc 
\faketableofcontents 
\part{} 


\section{Introduction}\label{introduction}

Unsupervised representation learning through diffusion models has emerged as a prominent area of research in computer vision. Several methods based on diffusion models \citep{zhang2022unsupervised, preechakul2022diffusion, abstreiter2021diffusion, baranchuk2021label} have been proposed for effective representation learning. Notably, \cite{baranchuk2021label} have  demonstrated the value of intermediate activations obtained from denoising networks, as they contain valuable semantic information that can be utilized for tasks like image representation and semantic segmentation. Their findings emphasize the effectiveness of diffusion models in learning meaningful visual representations. More recently , \cite{choi2022perception} have revealed that the restoration of data corrupted with specific noise levels provides an appropriate pretext task for the model to learn intricate visual concepts, and prioritizing such noise levels over other levels during training improves the performance of diffusion models. 


Despite the growing research on diffusion models in computer vision, there is still a notable lack of studies investigating the application of diffusion models to graph learning. Previous works, such as those by  \cite{haefeli2022diffusion} and \cite{jo2022score}, have primarily focused on utilizing diffusion models for the generation of discrete graph structures. However, the field of graph representation learning, which is a fundamental and challenging task in graph learning, has not yet exploited the potential of diffusion models. To successfully adapt and integrate diffusion models into graph representation learning and facilitate progress in this field, it is crucial to identify and comprehend the factors that impede the application of diffusion models.

\begin{figure}[t]
    \begin{subfigure}[b]{0.33\textwidth}
         \centering
         \includegraphics[width=\textwidth]{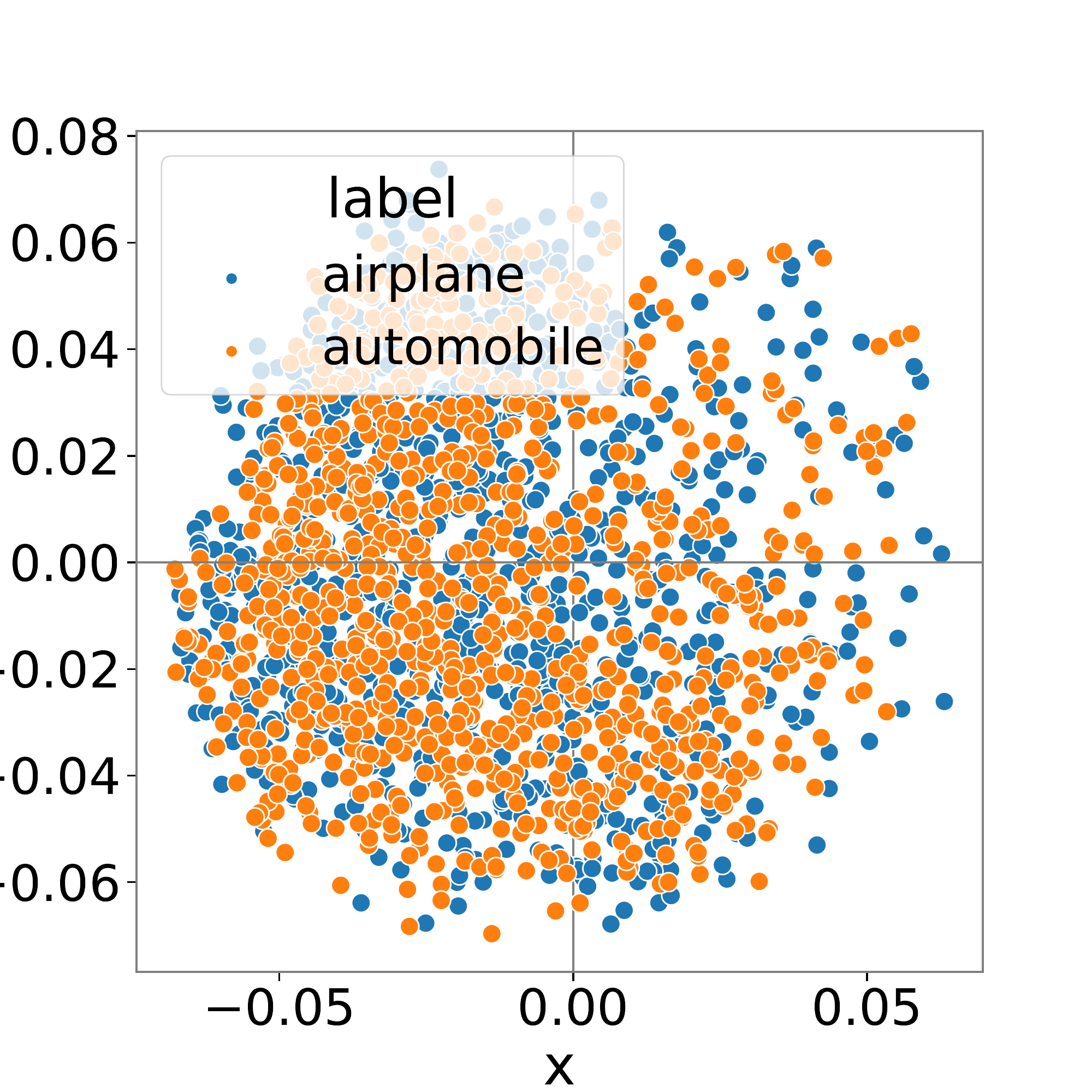}
         \caption{CIFAR-10}
         \label{cifar_dist}
     \end{subfigure}
     \begin{subfigure}[b]{0.33\textwidth}
         \centering
         \includegraphics[width=\textwidth]{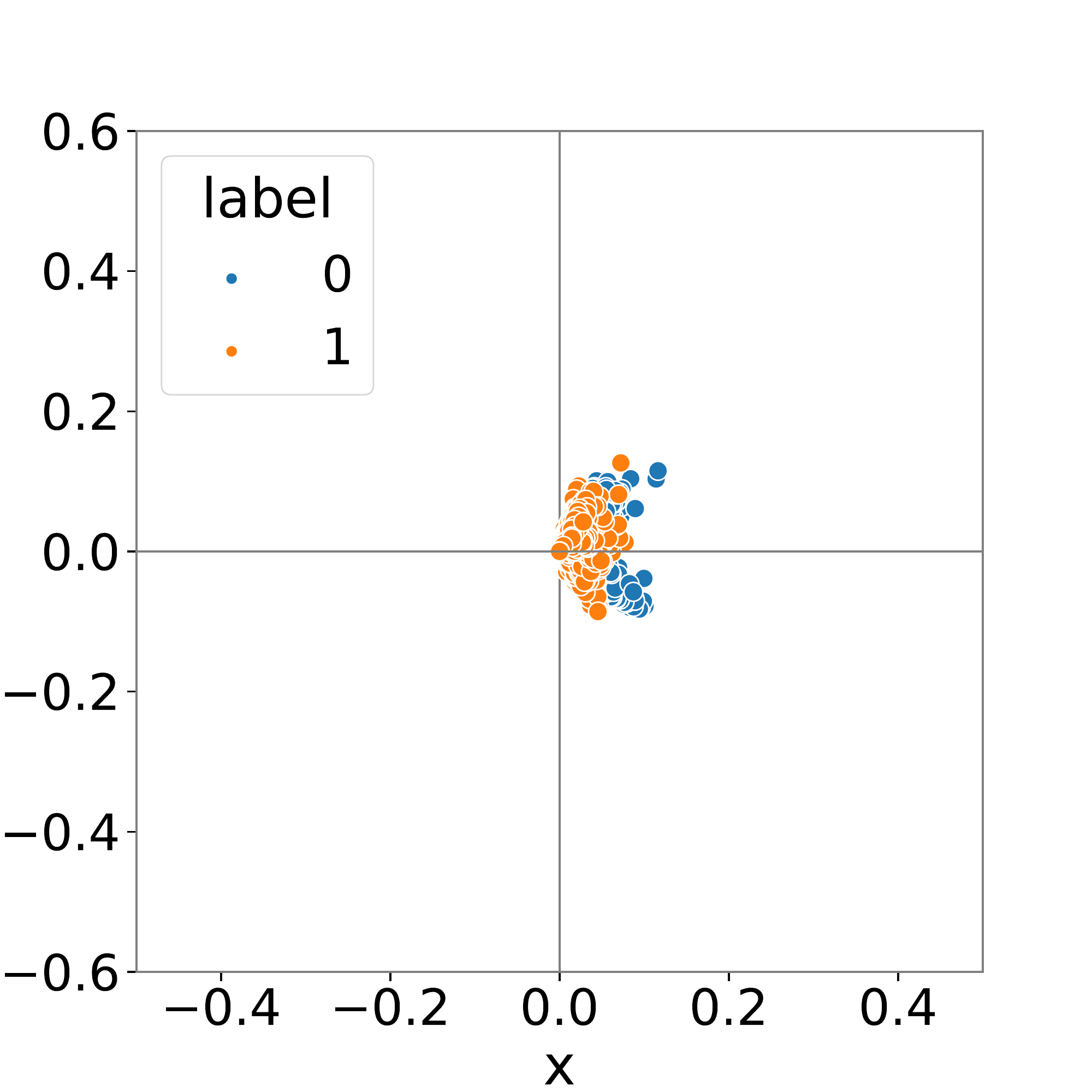}
         \caption{Amazon-Photo}
         \label{photo_dist}
     \end{subfigure}
     \begin{subfigure}[b]{0.33\textwidth}
         \centering
         \includegraphics[width=\textwidth]{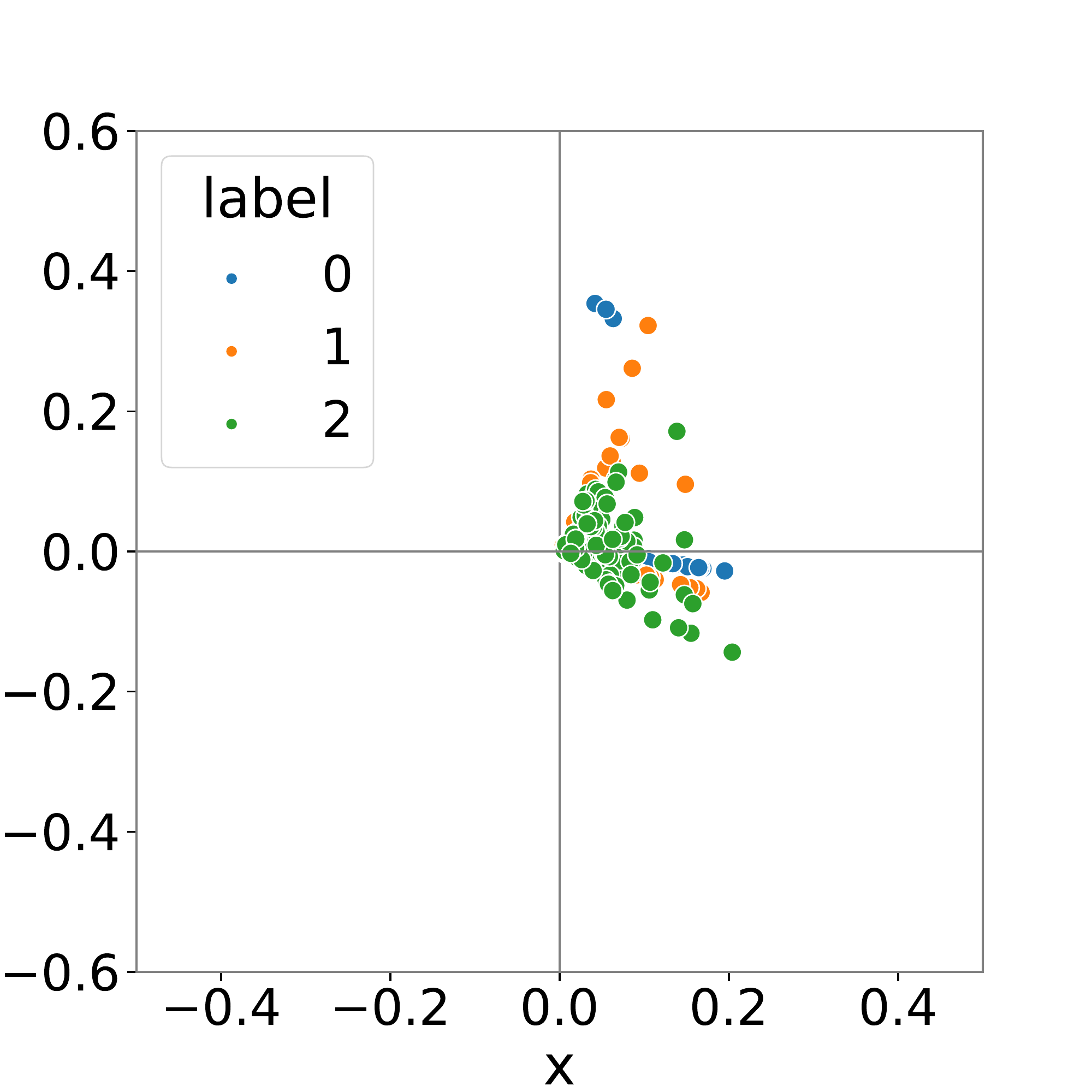}
         \caption{IMDB-M}
         \label{IMDBM_dist}
     \end{subfigure}
    \caption{2D visualization of SVD decomposition (a) Visualization of CIFAR-10's node features, different colors indicate different labels. (b) Visualization of two 
 classes of samples if Amazon-Photo. (c) Visualization of IMDB-M's graph features.}\label{distribution}
\end{figure}

To gain insights into the limitations of the vanilla diffusion models initially designed for image generation \citep{ho2020denoising}, we conduct experiments to investigate the underlying structural differences between images and graphs. Specifically, we employ singular value decomposition (SVD) on both image and graph data, and visualize the resulting data projections in a 2-dimensional plane, as shown in Figure \ref{distribution}. The figure illustrates that the projected data points from Amazon-Photo and IMDB-M exhibit strong anisotropic structures along only a few directions, while the projected images from CIFAR-10 form a relatively more isotropic distribution within a circular shape around the origin. This observation suggests that graph data may possess distinct anisotropic and directional structures that are less prominent in image data.  As we will demonstrate later, standard diffusion models with isotropic forward diffusion process will cause the inherrent signal-to-noise ratios (SNRs) to decline rapidly, making the standard diffusion models less effective in learning the anisotrpoic structures.  Therefore, it is imperative to develop  new approaches that can effectively account for these anisotropic structures. 

In this paper, we introduce directional diffusion models as a solution to account for the anisotropic structures, which can effectively  mitigate the rapid decline of signal-to-noise ratios issue.   Our approach involves incorporating data-dependent and directional noise in the forward diffusion process. We demonstrate that the intermediate activations obtained from the denoising network  effectively capture useful semantic and topological information required for downstream tasks. As a result, the proposed directional diffusion models offer a promising generative approach for graph representation learning. In our experimental evaluation, we conduct numerical experiments on 12 benchmark datasets encompassing both node and graph classification tasks. The results showcase the superior performance of our models compared to state-of-the-art contrastive learning and generative approaches \citep{hou2022graphmae}. Notably, for graph classification problems, our directional diffusion models even outperform supervised baselines, underscoring the immense potential of diffusion models in the field of graph learning.

Our  main contributions are as follows. 
\begin{enumerate}
\item  We contribute to the exploration of anisotropic structures in graph data, being among the pioneers in the literature. We demonstrate that the standard forward diffusion process with isotropic white noise leads to a rapid decline in signal-to-noise ratios for graph learning problems. This issue hampers the ability of denoising networks to extract fine-grained feature representations across a wide range of SNRs.

\item We propose novel diffusion models specifically designed for graph data, incorporating data-dependent and directional noise in the forward diffusion process. Our proposed models effectively address the issue of the rapid decline of SNRs, enabling better graph representation learning.  

\item Numerically, our proposed directional diffusion models outperform state-of-the-art self-supervised methods and even supervised methods on 12 benchmark datasets. Additionally, we provide comprehensive ablation studies to gain a deeper understanding of the mechanisms underlying directional diffusion models.
\end{enumerate}

\section{Related work}

\paragraph{Graph representation learning}
Graph representation learning aims to embed nodes or entire graphs into a low-dimensional vector space, where the structural and relational properties can be used for downstream tasks. Two prevalent paradigms for graph representation learning are contrastive learning and generative self-supervised learning. Contrastive learning approaches such as DGI \citep{velickovic2019deep}, Infograph \citep{sun2019infograph}, GraphCL \citep{you2020graph}, GRACE \citep{zhu2020deep}, and GCC \citep{qiu2020gcc}, have achieved promising results in some particular graph learning tasks. These methods leverage local-global mutual information maximization to develop unsupervised learning schemes for node and graph representation learning. GraphCL learns node embeddings that are invariant to graph-level transformations, while GRACE and GCC use subgraph sampling and graph perturbation to create augmented pairs. Generative self-supervised learning aims to recover missing parts of the input data through approaches such as GraphMAE \citep{hou2022graphmae}, a masked graph autoencoder that focuses on feature reconstruction by utilizing a masking strategy and scaled cosine error. This method outperforms both contrastive and masked state-of-the-art baselines, and it revitalizes the concept of generative self-supervised learning on graphs. GPT-GNN \citep{hu2020gpt} is a recent approach that leverages graph generation as the training objective. 

\paragraph{Denoising diffusion probabilistic models}

Denoising diffusion probabilistic models \citep{ho2020denoising, song2020denoising}, or simply diffusion models, are a class of probabilistic generative models that turn noise to a representative data sample and thus are mainly used for generation tasks \citep{dhariwal2021diffusion, rombach2022high}.
Recently, diffusion models have been used as a representation learning toolbox for computer vision problems \citep{preechakul2022diffusion, abstreiter2021diffusion, baranchuk2021label}.
For instance, \cite{preechakul2022diffusion} proposed Diff-AE, a method that jointly trains an encoder to discover meaningful feature representations from images and a conditional diffusion model that uses the representations as input conditions.  \cite{abstreiter2021diffusion} demonstrated that such an additional encoder can learn an infinite-dimensional latent code that achieves improvements in semi-supervised image classification tasks. Recently, diffusion models have also been used for processing graph data. \cite{haefeli2022diffusion} added noise to the adjacency matrix by a stochastic matrix to apply the diffusion model to graphs.  \cite{jo2022score} proposed the graph diffusion using stochastic differential equations (GDSS), which uses a system of stochastic differential equations in both graph structures and features for the graph generation.  However, this paper doesn't study the ability of graph representation with GDSS. To the best of our knowledge, there have been no works for diffusion-model-based graph representation learning.

\section{The effect of anisotropic structures}\label{de}


As mentioned in the introduction section, there are notable disparities in structural properties between graphs and natural images. In addition to the Amazon-Photo and IMDB-M datasets, we conducted similar analyses on all the other graph benchmark datasets described in Section \ref{exp}. The additional results can be found in the appendix. Furthermore, it is important to highlight that these anisotropic structures, often referred to as categorical directional dependence, are also commonly observed in natural language data \citep{gao2019representation, li2020sentence}. It is interesting to note that diffusion models have not yet achieved significant success in the field of natural language processing, which further emphasizes the importance of exploring and addressing the challenges posed by anisotropic structures in the context of diffusion models.



This section delves further into the study of how the anisotropic structures of graphs hinder the effectiveness of vanilla diffusion models in graph learning problems. In the vanilla forward diffusion process of diffusion models, isotropic Gaussian noise is sequentially added to the data point $x_{0} \sim q(x_{0})$ until it becomes white noise following $\mathcal{N}\left(0, \mathbf{I}\right)$\footnote{Bold symbols are used for matrices, but not for vectors.}. This process is reasonable when the data follow isotropic distributions, as it gradually transforms the data point into noise and generates noisy data points with a wide range of SNRs. However, in the case of anisotropic data distributions, adding isotropic noise can rapidly contaminate the data structure, causing the SNRs to decline rapidly towards zero. Consequently, denoising networks become unable to extract fine-grained feature representations at different SNR scales.

To investigate the impact of adding isotropic noise on learning anisotropic graphs, we design an experiment to measure the SNRs for both node and graph classification tasks in a linear-separable hidden space at each forward step and observe how these SNRs change along the forward diffusion process. First, we pre-train a graph neural network (GNN) denoted as $\mathbf{E}$ to serve as a feature extractor that projects the graph data into a linear-separable hidden space. Then, we optimize the weight vector $w\in\mathbb{R}^{d\times 1}$ in the hidden space using Fisher's linear discriminant analysis. The weight vector $w$ is employed to calculate the ${\rm SNR} = {w^\top S_B w}/{w^\top S_W w}$  at each forward diffusion step, where $S_B$ is the scatter between class variability and $S_W$ is the scatter within-class variability. This SNR quantifies the discriminative power of the learned representations at different steps of the diffusion process.

\begin{figure}[t]
     \centering
     \begin{subfigure}[b]{.45\textwidth}
         \centering
         \includegraphics[width=\textwidth]{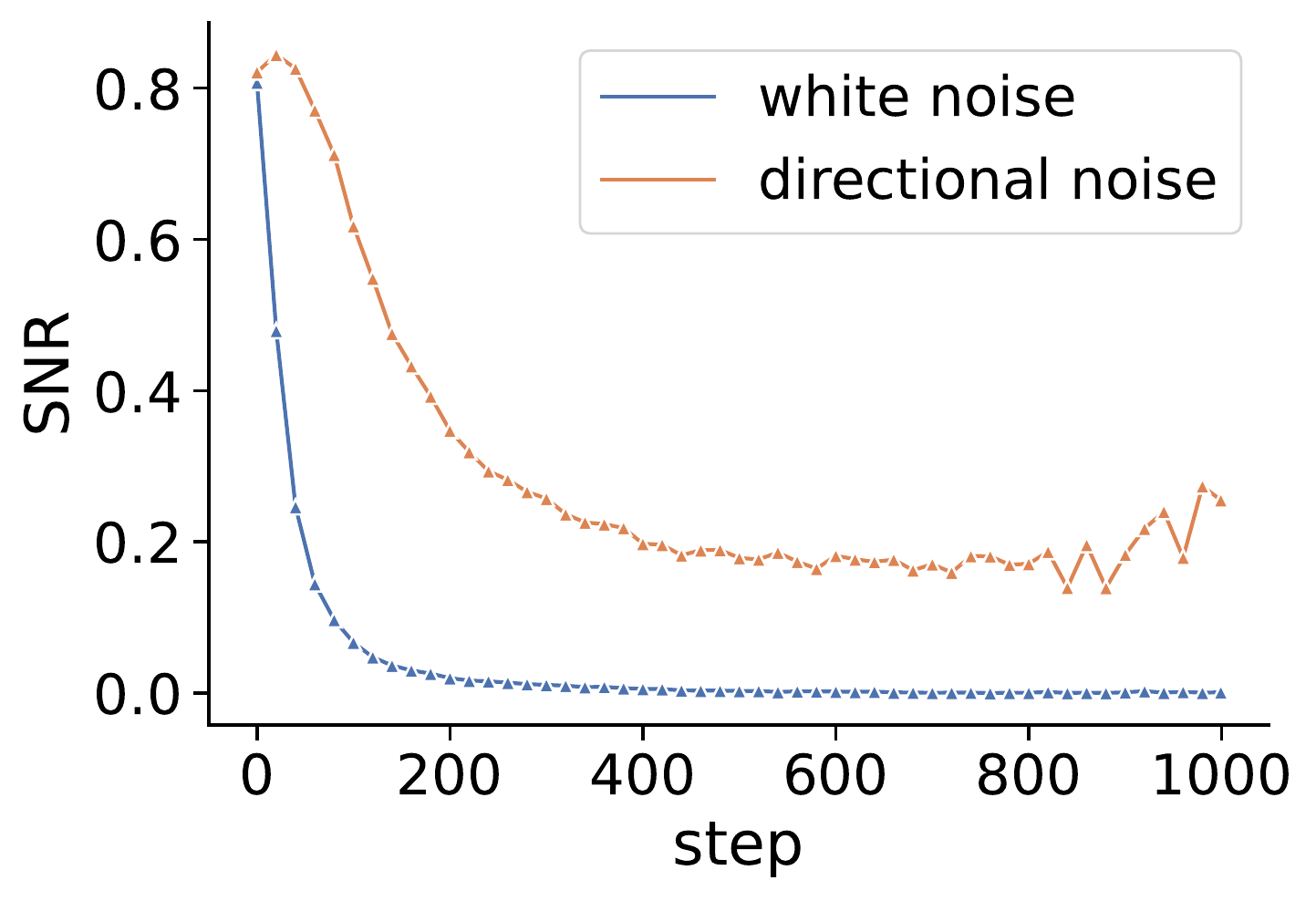}
         \caption{Amazon-Photo}
         \label{white_dist}
     \end{subfigure}
     \begin{subfigure}[b]{.45\textwidth}
         \centering
         \includegraphics[width=\textwidth]{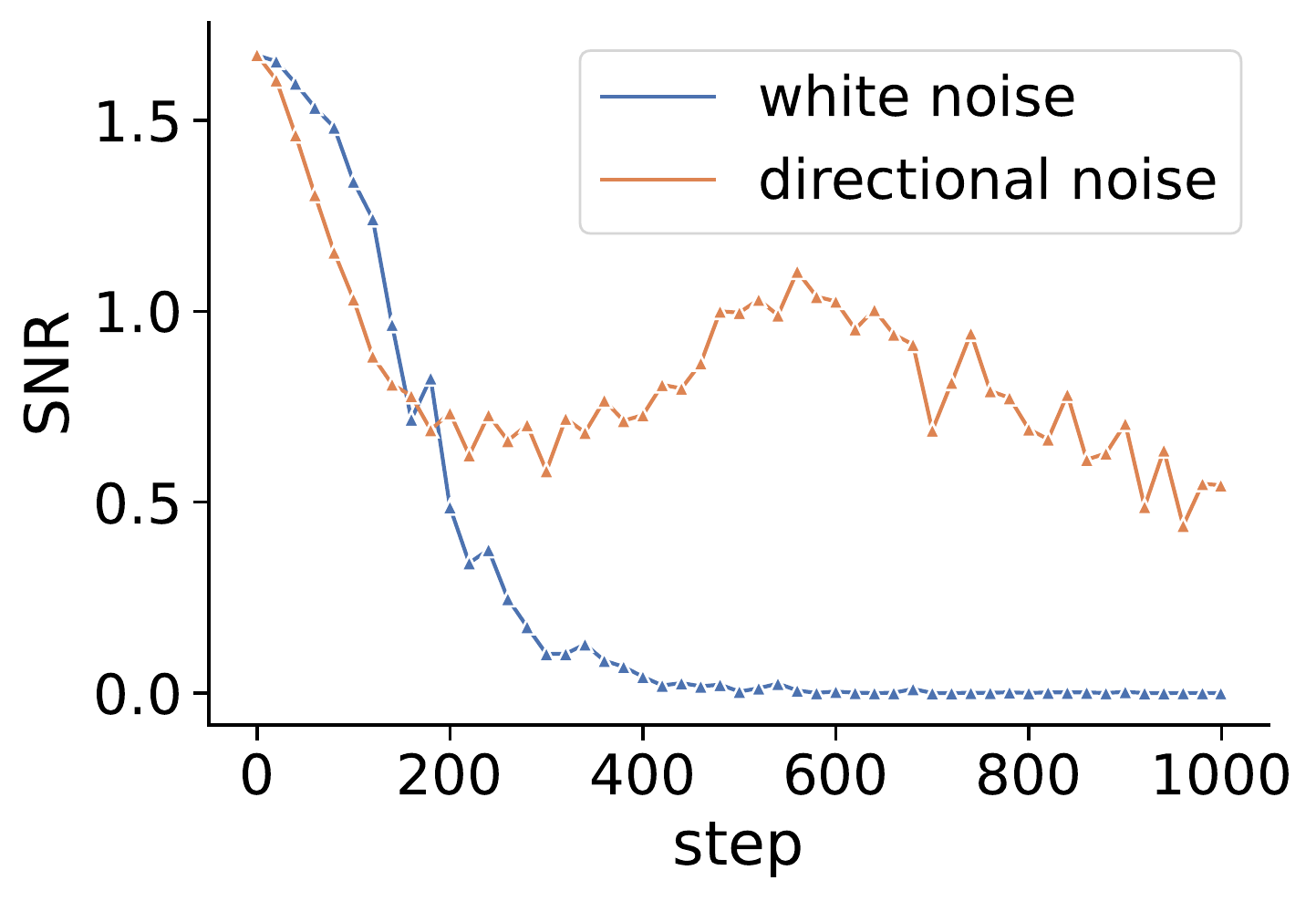}
         \caption{IMDB-M}
         \label{ddm_dist}
     \end{subfigure}
    \caption{The signal-to-noise ratio curve along different diffusion steps. }
    \label{distance}
\end{figure}

We conducted this experiment on all graph benchmark datasets to assess the impact of isotropic noise on learning anisotropic graphs. Here, we present the results for IMDB-M and Amazon-Photo in Figure \ref{white_dist}, while the additional results can be found in the appendix. In Figure \ref{white_dist}, we observe that for anisotropic graph data and isotropic noise, the SNR rapidly decreases to 0 at around 50 steps for Amazon-Photo and 400 steps for IMDB-M.  Furthermore, the SNR remains close to 0 thereafter, indicating that the incremental isotropic white noise quickly obscures the underlying anisotropic structures or signals. Consequently, the denoising networks are unable to learn meaningful and discriminative feature representations that can be effectively utilized for downstream classification tasks. In contrast, when utilizing our directional diffusion models, which incorporate a data-dependent and directional forward diffusion process (to be introduced later), the SNR declines at a slower pace. This slower decline enables the extraction of fine-grained feature representations with varying SNRs, preserving the essential information of the anisotropic structures.

Overall, these studies underscore the significance of considering anisotropic data structures when designing forward diffusion processes and the corresponding diffusion models, especially in the context of graph data where anisotropic structures are commonly observed.


\section{Directional diffusion models}\label{ddm}

In this section, we will begin by introducing the necessary notation for our discussion. We then propose the directional diffusion models (DDMs) as an extension of the vanilla diffusion models, specifically designed for graph representation learning. 
We will also discuss how to extract feature representations from DDMs, which is crucial for downstream tasks.

\paragraph{Notation}

We denote a graph by $G = (\mathbf{V}, \mathbf{A}, \mathbf{X})$, where $\mathbf{V}$ is the node set, $N = |\mathbf{V}|$ is the node number,  and $\mathbf{A} \in \mathbb{R}^{N \times N}$ is the adjacency matrix (binary or weighted). $\mathbf{X} = \{x_{1}, x_{2}, \cdots, x_{N}\}\in \mathbb{R}^{N \times d}$ is the node feature matrix. Our goal is to learn a network, $f: \mathbb{R}^{N \times d} \times \mathbb{R}^{N \times N} \to \mathbb{R}^{N \times d_{h}}$ to encode graph features into representations $\mathbf{H} = \{h_{1}, h_{2}, \cdots, h_{N}\} \in \mathbb{R}^{N \times d_{h}}$, and $h_{i} \in \mathbb{R}^{ d_{h}}$ is the representation for each node $i$, denoted as $\mathbf{H} = f (\mathbf{X}, \mathbf{A})$. These representations can be used for downstream tasks, such as graph classification and node classification.

\paragraph{Directional diffusion models}

In the previous section, our investigations unveiled a critical factor responsible for the subpar performance of vanilla diffusion models in graph learning: the rapid decline of signal-to-noise ratios. To address this challenge, we propose a solution called \textit{directional noise}, which involves transforming the initially isotropic Gaussian noise into an anisotropic noise by incorporating two supplementary constraints. These two constraints play a crucial role in improving the vanilla diffusion models. 


Let $G_t = (\mathbf{A}, \mathbf{X}_t)$ be the working solution  of the $t$-th forward  diffusion step, where $\mathbf{X}_t = \{x_{t,1}, x_{t,2}, ..., x_{t, N}\}$ represents the learned features at the $t$-th step. To be specific, the node feature $x_{t, i} \in \mathcal{R}^d $ of node $i$ at time  $t$ is obtained as
\begin{align}
x_{t, i} &= \sqrt{\Bar{\alpha}_{t}}x_{0, i} + \sqrt{1-\Bar{\alpha}_{t}}\epsilon^{\prime},  \label{dn}  \\
\epsilon^{\prime} &= {\rm sgn}(x_{0, i}) \odot  | \bar{\epsilon} |,  \label{st1} \\
\bar{\epsilon} &= \mu + \sigma\odot\epsilon \quad {\rm where} \ \epsilon \sim \mathcal{N}\left(0, \mathbf{I}\right),  \label{st2}
\end{align}
where $x_{0, i}$ is the raw feature vector of node $i$,  $\mu \in \mathcal{R}^d$ and  $\sigma \in \mathcal{R}^d$ consists of the means and standard deviations of the $d$ features across $N$ nodes respectively, and $\odot$ denotes the Hadamard product. During the mini-batch training, $\mu$ and $\sigma$ are calculated using graphs within the batch. The parameter $\Bar{\alpha}_{t}:= \prod_{i=0}^{t}(1-\beta_{i})  \in (0,1)$ represents the fixed variance schedule \citep{ho2020denoising} and is parameterized by a decreasing sequence $\{\beta_{1: T} \in (0, 1)\}$. 

\begin{figure}[t]
    \centering
    \includegraphics[width=\textwidth]{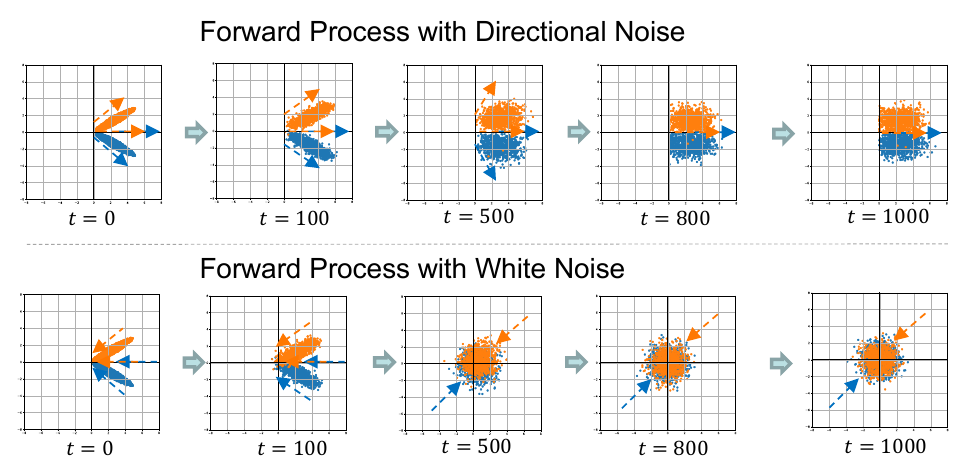}
    \caption{Directional noise vs white noise. We sample directional and white noises\protect\footnotemark[1],  and then iterative add these noises  in the same scheme. The upper panel collects the samples with directional noises at the diffusion steps $t=0, 100, 500, 800, 1000$, and the lower one shows the sample with white noises at the same diffusion steps. The two different colors indicate two different classes. }
    \label{noise com}
\end{figure}
\footnotetext[1]{The generation parameters are provided in the appendix.}

Compared to the vanilla forward diffusion process, our directional diffusion models incorporate two additional constraints, namely \eqref{st1} and \eqref{st2}. The second constraint, \eqref{st2}, transforms the data-independent Gaussian noise into an anisotropic and batch-dependent noise. In this constraint, each coordinate of the noise vector shares the same empirical mean and empirical standard deviation as the corresponding coordinate in the data within the same batch. This constraint restricts the diffusion process to the local neighborhood of the batch, preventing excessive deviation from the batch and maintaining local coherence. 
The first constraint \eqref{st1} introduces as an angular direction that rotates the noise $\bar{\epsilon}$ into the same hyperplane of the  feature $x_{0,i}$, ensuring that adding noise will not cause the noisy features to be in the opposite direction of $x_{0,i}$. By preserving the directionality of the original feature, this constraint helps maintain the inherent data structure during the forward diffusion process.  These two constraints work in tandem to ensure that the forward diffusion process respects the underlying data structure and prevents the rapid washing away of signals. As a result, the SNR decays slowly, allowing our directional diffusion models to effectively extract meaningful feature representations at various SNR scales. This, in turn, benefits downstream tasks by providing more reliable and informative representations. 


To illustrate the impact of directional noise, we refer to the experiments conducted in Section \ref{de}. Our newly proposed "directional noise" ensures a smoother decline of the signal-to-noise ratios (SNRs) throughout the diffusion process, which confirms our initial intuition. In order to further visualize the differences between using directional noise and isotropic white noise in the diffusion process, we conducted simulations on two ellipses and sequentially added noise, as depicted in Figure \ref{noise com}. The figure clearly illustrates the distinct behaviors exhibited by the two types of noise. With directional noise, the samples maintain a clear decision boundary, indicating the preservation of discriminative structures during the diffusion process. Conversely, samples with white noise quickly blend into pure noise, leading to the loss of meaningful information. This visual comparison clearly highlights the superiority of directional noise in preserving the structural information of the data during the diffusion process.

\paragraph{Model architecture}

We follow the same training strategy as in the vanilla diffusion models, where we train a denoising network $f_{\theta}$ to approximate the reverse diffusion process.  
Since the posterior of the forward process with directional noise cannot be expressed in a closed form, we borrow the idea from \cite{bansal2022cold, li2022diffusion} and let the denoising model $f_{\theta}$ directly predict $\mathbf{X}_0$.
The loss function $\mathcal{L}$ is defined as the expected value of the Euclidean distance between the predicted feature representation $f_{\theta}(\mathbf{X_{t}}, \mathbf{A}, t)$ and the original feature representation $\mathbf{X_{0}}$:
\begin{equation}\label{loss}
    \mathcal{L} = \mathbb{E}_{\mathbf{X_0}, t} \| f_{\theta}(\mathbf{X_{t}}, \mathbf{A}, t)  - \mathbf{X_{0}} \|^{2}. 
\end{equation}
This loss function ensures that the model predicts $\mathbf{X}_0$ at every step.

\begin{figure}[t]
\centering
\includegraphics[width=\textwidth]{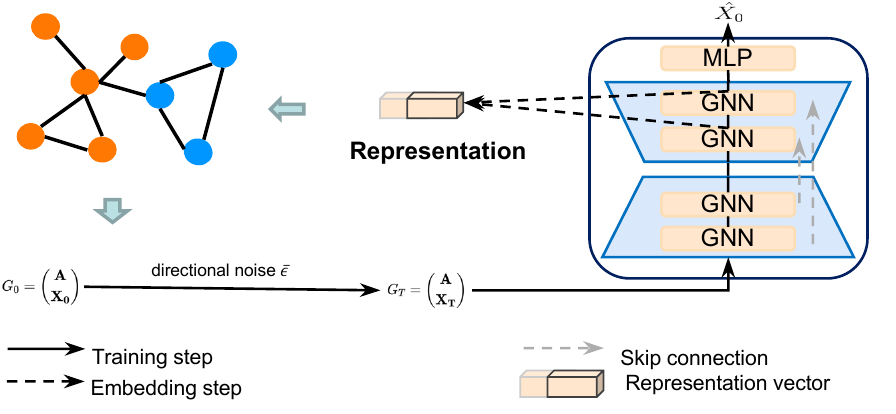}
\caption{The pipeline of our model: (1) Adding directional noise to the original graph $G_0$. (2) Extracting the last two GNN's feature maps in the denoising network as the representation of the graph.}

\label{framework}
\end{figure}

To parameterize the denoising network $f_{\theta}$, we utilize a symmetrical architecture that incorporates Graph Neural Networks (GNNs), taking inspiration from the successful UNet architecture in computer vision \citep{dhariwal2021diffusion}. Figure~\ref{framework} provides an illustration of our DDM framework, which consists of four GNN layers and one multilayer perception (MLP). The first two GNN layers serve as the {\it encoder}, responsible for denoising the target node by aggregating neighboring information. The last two GNN layers function as the {\it decoder}, mapping the denoised node features to a latent code and smoothing the latent code between neighboring nodes. To address the potential issue of over-smoothing and account for long-distance dependencies on the graph, we introduce skip-connections between the encoder and the decoder within our architecture. The MLP architecture within the DDM transforms the latent code into the original feature matrix $\mathbf{X}_0$, allowing the latent code to contain meaningful compressed knowledge while preventing the decoder from degrading into an identical mapping. The algorithm, along with the mini-batch training procedure, can be found in the appendix.

\paragraph{Learning representations} 

For a given graph $G = (\mathbf{A}, \mathbf{X})$, the learned node-level representations are obtained from the activations of the denoising network $f_{\theta}$ at user-selected time steps. It is important to note that we only utilize the activations from the decoder of $f_{\theta}$ since they incorporate the encoder activations through skip connections. As depicted in Figure~\ref{framework}, at each time step $k$, we introduce $k$ steps of directional noise following ~\eqref{dn} and employ the denoising network $f_{\theta}$ to denoise and compress the noisy data $\mathbf{X}_k$. The decoder of $f_{\theta}$ maps the denoised node features to a latent code while smoothing the latent code among neighboring nodes. We extract the activations from the decoder of $f_{\theta}$ and concatenate them to obtain $\mathbf{H}_{k} = \{h_{k, 1}, h_{k, 2}, \cdots, h_{k, N}\} \in \mathbb{R}^{N \times d_{h}}$. The complete pipeline is presented in the appendix.

\section{Experiments}\label{exp}

This section provides an evaluation of the directional diffusion models from two perspectives. Firstly, we compare our models with existing state-of-the-art methods on various graph learning tasks, including node and graph classification tasks. This allows us to assess the effectiveness of our approach for graph representation learning problems. Secondly, we conduct several studies to gain a better understanding of the effect of our directional noise and evaluate the necessity of our design choices.

In all experiments, we follow a two-step process. First, we pretrain a DDM on the dataset in an unsupervised manner. Then, we extract feature representations from diffusion steps ${50, 100, 200}$  using the pretrained model. Although this approach is inspired by the experimental results and insights from Section \ref{de}, it is deliberately not fine-tuned for each dataset. Ideally, fine-tuning with carefully selected steps for each dataset could further improve the performance.

\subsection{Graph classification}\label{exp:gc}

To demonstrate the effectiveness of our method, we  compare with state-of-the-art (SOTA) unsupervised learning methods, including GCC\cite{qiu2020gcc}, Infograph \cite{sun2019infograph}, GraphCL \cite{you2020graph}, JOAO \cite{you2021graph}, MVGRL \cite{hassani2020contrastive}, and GraphMAE\cite{hou2022graphmae}. We also compare with  supervised learning methods, including GIN \cite{xu2018powerful} and DiffPool \cite{ying2018hierarchical}. The experiments are carried out on seven widely-used datasets, namely MUTAG, IMDB-B, IMDB-M, PROTEINS, COLLAB, and REDDIT-B \cite{yanardag2015deep}. Node degrees are used as initial node features for IMDB-B, IMDB-M, REDDIT-B, and COLLAB, while node labels are used for MUTAG and PROTEINS, consistent with the previous literature  \citep{hou2022graphmae}. We extracted the graph-level representations at different steps and trained and tested independent LIBSVM \cite{chang2011libsvm} on them for evaluation. The final prediction was obtained by majority vote, and we reported the mean 10-fold cross-validation accuracy with standard deviation after five runs. We leave details on hyper-parameters can be found in the appendix.

Table~\ref{graph-exp} presents the results, demonstrating that our DDM achieves the best or competitive performance across all benchmark datasets. Particularly noteworthy is that our DDM surpasses even the supervised approaches in certain experiments, such as IMDB-B, COLLAB, and MUTAG. This exceptional performance can be attributed to two factors. First, from a data perspective, the node features in these datasets contain limited information, which can hinder the accuracy of supervised learning \citep{hou2022graphmae}. By utilizing the directional noise diffusion, our DDM acts as a pseudo-infinite-step data augmentation technique that generates numerous samples while preserving the classification boundary. This augmentation improves the effectiveness of unsupervised learning. Second, from a model perspective, the DDM framework leverages the power of directional noise and ensures that the learned representations capture meaningful information by avoiding the rapid decay of signal-to-noise ratios. 

\begin{table}[t]
\centering
\caption{Results in unsupervised representation learning for graph classification. }
\label{graph-exp}
\renewcommand\arraystretch{1.2}
\resizebox{\columnwidth}{!}
{
\begin{tabular}{c|cccccc}
\Xhline{1.2pt}
Dataset   & IMDB-B      & IMDB-M      & COLLAB     & REDDIT-B   & PROTEINS    & MUTAG       \\ \hline
GIN       & 75.1±5.1    & 52.3±2.8    & 80.2±1.9   & 92.4±2.5   & 76.2±2.8    & 89.4±5.6    \\
DiffPool  & 72.6±3.9    & -           & 78.9±2.3   & 92.1±2.6   & 75.1±2.3    & 85.0±10.3   \\ \hline
Infograph & 73.03±0.87  & 49.69±0.53  & 70.65±1.13 & 82.50±1.42 & 74.44±0.31  & 89.01±1.13  \\
GraphCL   & 71.14±0.44  & 48.58±0.67  & 71.36±1.15 & 89.53±0.84 & 74.39±0.45  & 86.80±1.34  \\
JOAO      & 70.21±3.08  & 49.20±0.77  & 69.50±0.36 & 85.29±1.35 & 74.55±0.41  & 87.35±1.02  \\
GCC       & 72          & 49.4        & 78.9       & 89.8       & -           & -           \\
MVGRL     & 74.20±0.70  & 51.20±0.50  & -          & 84.50±0.60 & -           & 89.70±1.10  \\
GraphMAE  & 75.52±0.66  & 51.63±0.52  & 80.32±0.46 & 88.01±0.19 & 75.30±0.39  & 88.19±1.26  \\ \hline
DDM       & \textbf{76.40±0.22} & \textbf{52.53±0.31} & \textbf{81.72±0.31} & 89.15 ±1.3 & \textbf{75.47 ±0.50} & \textbf{91.51 ±1.45} \\ 
\Xhline{1.2pt}
\end{tabular}
}
\end{table}

\begin{table}[t]
\centering
\caption{Results in unsupervised representation learning for node classification.}
\label{tab:node-exp}
\renewcommand\arraystretch{1.2}
\resizebox{\columnwidth}{!}
{
\begin{tabular}{c|cccccc}
\Xhline{1.2pt}
Dataset  & Cora     & Citeseer    & PubMed    & Ogbn-arxiv & Computer              & \multicolumn{1}{c}{Photo} \\ \hline
GAT      & 83.0 ± 0.7 & 72.5 ± 0.7    & 79.0 ± 0.3  & 72.10 ± 0.13 & 86.93 ± 0.29            & 92.56 ± 0.35              \\ \hline
DGI      & 82.3 ± 0.6 & 71.8 ± 0.7    & 76.8 ± 0.6  & 70.34 ± 0.16 & 83.95 ± 0.47          & 91.61 ± 0.22              \\
MVGRL    & 83.5 ± 0.4 & 73.3 ± 0.5    & 80.1 ± 0.7  & -          & 87.52 ± 0.11          & 91.74 ± 0.07              \\
BGRL     & 82.7 ± 0.6 & 71.1 ± 0.8    & 79.6 ± 0.5  & 71.64 ± 0.12 & 89.68 ± 0.31          & 92.87 ± 0.27              \\
InfoGCL  & 83.5 ± 0.3 & 73.5 ± 0.4    & 79.1 ± 0.2  & -          & \multicolumn{1}{c}{-} & \multicolumn{1}{c}{-}     \\
CCA-SSG  & 84.0 ± 0.4 & 73.1 ± 0.3    & 81.0 ± 0.4  & 71.24 ± 0.20 & 88.74 ± 0.28          & 93.14 ± 0.14              \\
GPT-GNN  & 80.1 ± 1.0 & 68.4 ± 1.6    & 76.3 ± 0.8  & -          & \multicolumn{1}{c}{-} & \multicolumn{1}{c}{-}     \\
GraphMAE & 84.2 ± 0.4 & 73.4 ± 0.4    & 81.1 ± 0.4  & 71.75 ± 0.17 & 88.63 ± 0.17         & 93.63 ± 0.22              \\ \hline
DDM      & 83.4 ± 0.2 & \textbf{74.3 ± 0.3}    & \textbf{81.7 ± 0.8}  & 71.29 ± 0.18 &\textbf{90.56 ± 0.21}            & \textbf{95.09 ± 0.18}                \\ \Xhline{1.2pt}
\end{tabular}
}
\end{table}

\subsection{Node classification}

To assess the quality of the node-level representations produced by our method, we evaluated the performance of DDM on six standard benchmark datasets: Cora, Citeseer, PubMed \citep{yang2016revisiting}, Ogbn-arxiv \citep{hu2020open}, Amazon-Computer \citep{zhang2021canonical}, and Amazon-Photo \citep{zhang2021canonical}. We followed the publicly available data-split schema and utilized the evaluation protocol used in previous approaches. Graph-level representations were extracted at different diffusion steps, and an independent linear classifier was trained for each step. The final prediction was obtained through majority voting, and we reported the mean accuracy on the test nodes. Details of the hyperparameters can be found in the appendix.

We compare DDM with state-of-the-art generative unsupervised models, namely GPT-GNN \citep{hu2020gpt} and GraphMAE \citep{hou2022graphmae}. Additionally, we include the results of contrastive unsupervised models for comparison, namely DGI \citep{velickovic2019deep}, MVGRL \citep{hassani2020contrastive}, GRACE \citep{zhu2020deep}, BGRL \citep{thakoor2021large}, InfoGCL \citep{xu2021infogcl}, and CCA-SSG \citep{zhang2021canonical}.  As shown in Table~\ref{tab:node-exp}, DDM achieves competitive results across all benchmark datasets. This indicates that the generative diffusion method is capable of learning meaningful node-level representations, and DDM is effective for node-level tasks. Notably, the node features used in node classification are text embeddings, highlighting the efficacy of our directional noise in continuous word vector spaces.

\subsection{Understanding the noise}

The above studies provide compelling evidence that our approach surpasses or is comparable to existing SOTA methods. In order to gain a deeper understanding,  we conduct a comprehensive investigation into the impact of different types of noise. Furthermore, we analyzed directional noise by removing the two constraints, specified as ~\eqref{st2} and ~\eqref{st1}, respectively, to examine their individual effects.

\begin{figure}[t]
     \centering
     \begin{subfigure}[b]{0.48\textwidth}
         \centering
         \includegraphics[width=\textwidth]{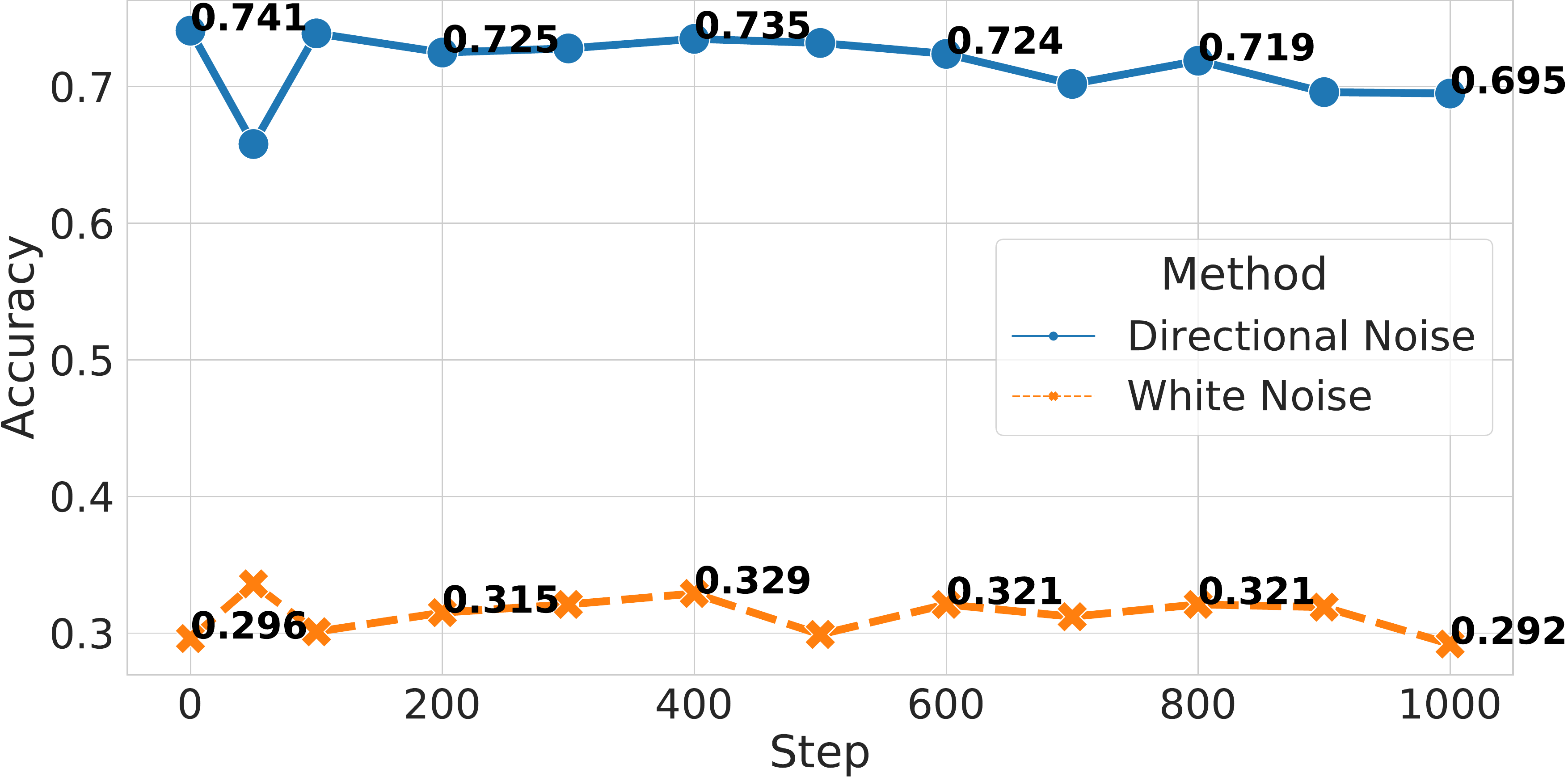}
         \caption{Citeseer}
     \end{subfigure}
      \begin{subfigure}[b]{0.48\textwidth}
         \centering
         \includegraphics[width=\textwidth]{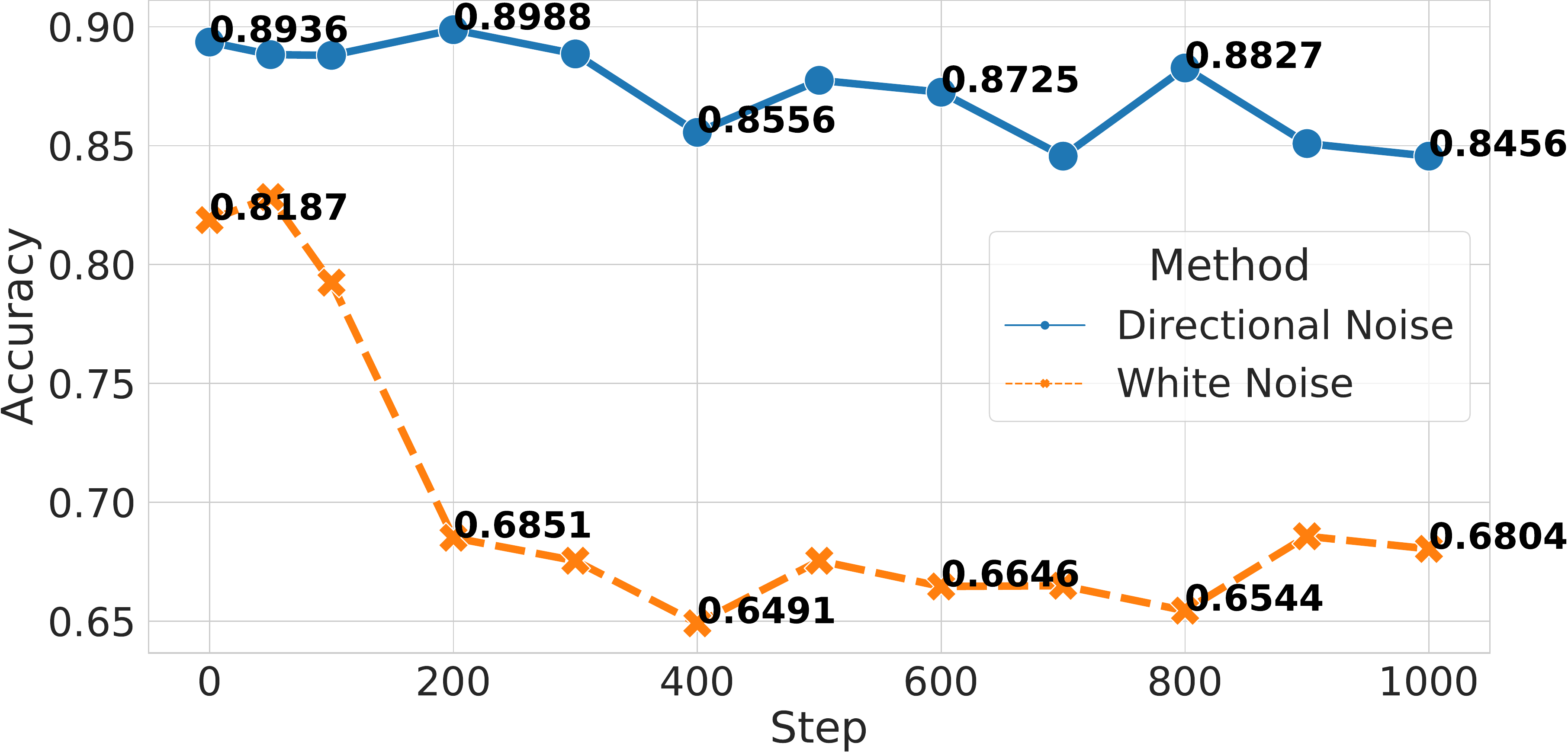}
         \caption{MUTAG}
     \end{subfigure}
     \quad
    \begin{subfigure}[b]{0.48\textwidth}
         \centering
         \includegraphics[width=\textwidth]{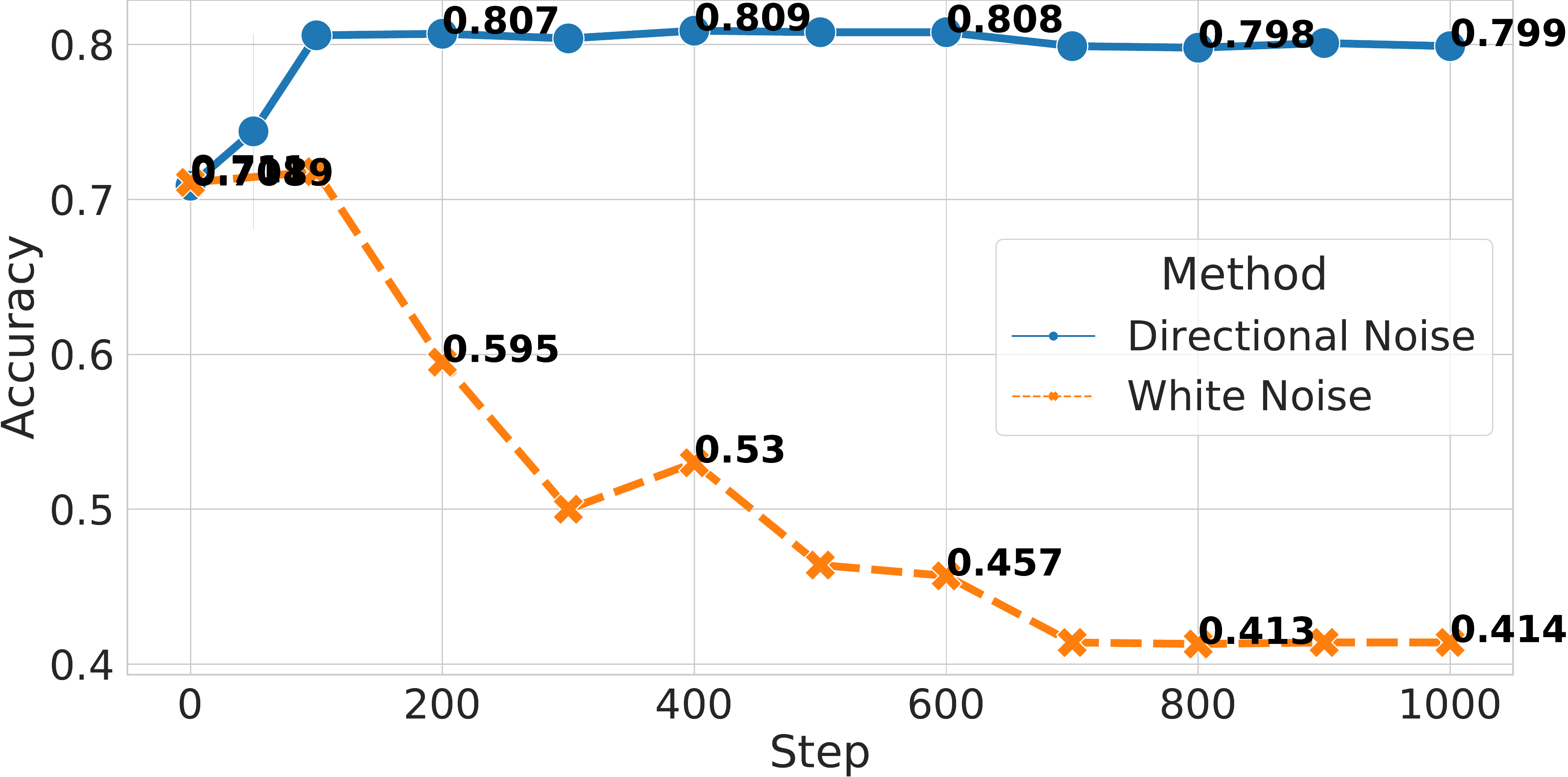}
         \caption{PubMed}
     \end{subfigure}
    \begin{subfigure}[b]{0.48\textwidth}
         \centering
         \includegraphics[width=\textwidth]{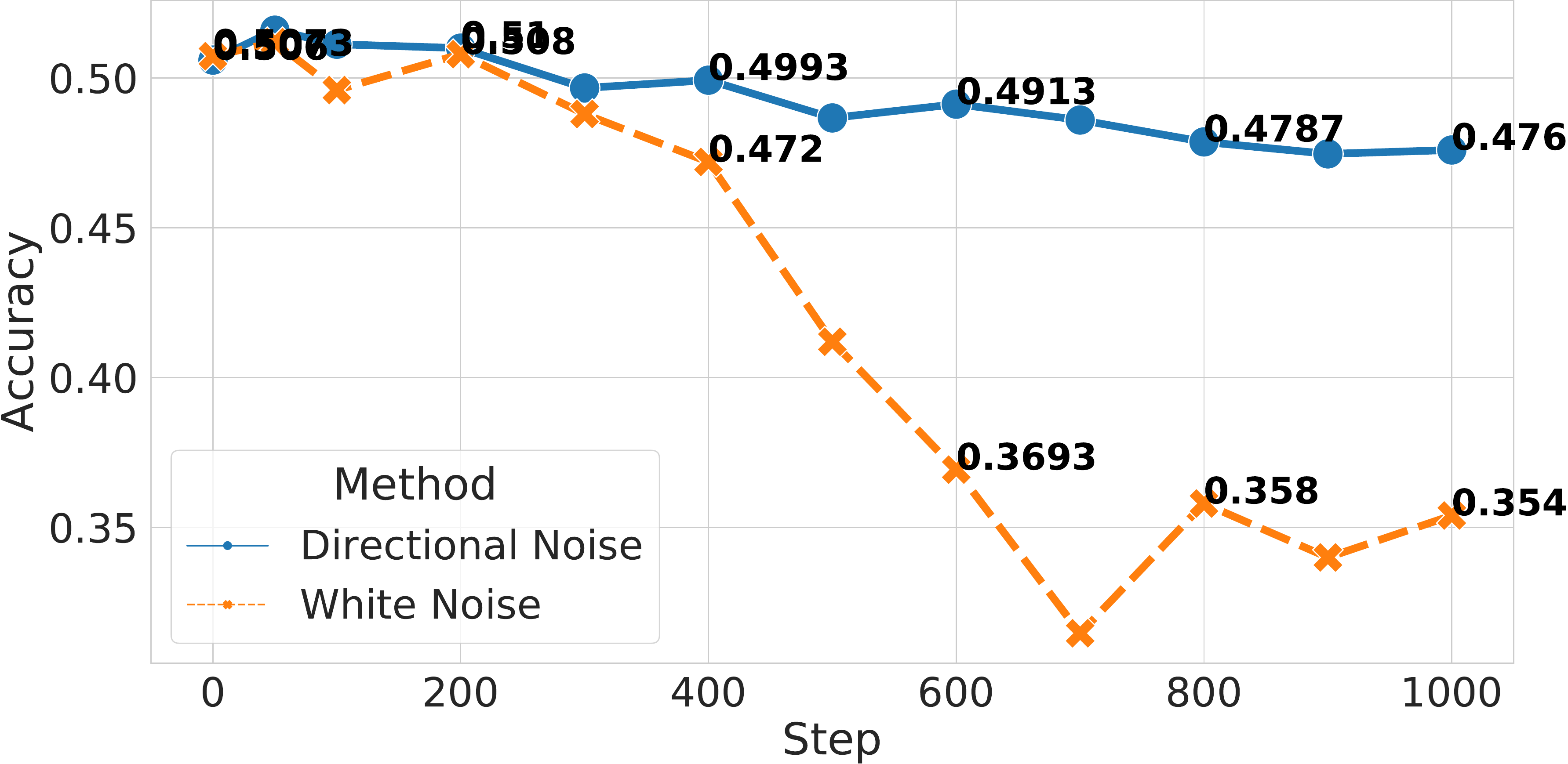}
         \caption{IMDB-M}
     \end{subfigure}
    \caption{The downstream tasks' accuracy of representations extracted from models trained with directional noise and white noise at every step of the reverse process.}
    \label{repre-step}
\end{figure}

We conduct a comparison of the representations extracted from models trained with directional noise and white noise at each step of the reverse diffusion process. The results, as presented in Figure ~\ref{repre-step}, reveal significant differences between the two approaches. With white noise, only the representations corresponding to the early steps of the reverse diffusion process contain useful topological information, while the majority of representations for later steps become uninformative. This stands in stark contrast to the case of directional noise, where the learned representations consistently preserve sufficient information for downstream classification tasks.

In our additional experiments, we observed that directional diffusion models consistently outperform vanilla diffusion models across all datasets, particularly in node classification tasks. This superior performance can be attributed to the nature of node classification datasets, which often utilize word vectors as node features. These word vectors exhibit higher feature dimensionality and greater anisotropy. The effectiveness of our directional approach is further supported by these findings, reinforcing its value in graph representation learning.

\begin{minipage}[hbt]{0.6\linewidth}

Lastly, we conduct an ablation study to examine the effects of the two key constraints. The results are presented in Table \ref{tab:ab study}, where "w/o R" indicates the removal of the constraint \eqref{st1} and "w/o S\&R" indicates the removal of both constraints. As shown in Table \ref{tab:ab study}, the introduction of anisotropic Gaussian noise generated through \eqref{st2} led to a significant improvement compared to isotropic Gaussian noise. Furthermore, the inclusion of constraint \eqref{st1} provided an additional and indispensable improvement. This finding further confirms the importance of making the noise in the forward diffusion process data-dependent and anisotropic. 

\end{minipage}
\begin{minipage}[hbt]{0.4\linewidth}
\centering
\captionof{table}{Ablation studies.}
\renewcommand\arraystretch{1.3}
\resizebox{0.9\columnwidth}{!}{%
\begin{tabular}{c|ccc}
\Xhline{1.2pt}
Dataset  & w/o S\&R & w/o R & Full       \\ \hline \hline
Citeseer & 34.37 & 60.77    & \textbf{74.30} \\
PubMed   & 73.03 & 77.60    & \textbf{81.23} \\ \hline \hline
IMDB-M   & 49.80  & 50.87    & \textbf{52.74} \\
COLLAB   & 80.50 & 81.04    & \textbf{81.72} \\
MUTAG    & 82.89 & 87.25    & \textbf{90.41} \\ 
\Xhline{1.2pt}
\end{tabular}
}
\label{tab:ab study}
\end{minipage}

\section{Conclusions}

This paper unveils the anisotropic structures present in graphs, which render vanilla diffusion models inadequate for graph representation learning. To address this limitation, we introduce directional diffusion models, a novel class of diffusion models that leverage data-dependent and anisotropic noise for unsupervised graph representation learning. Through experiments conducted on 12 benchmark datasets, we demonstrate the effectiveness of our proposed method.

There are several promising avenues for future research. One direction is to develop methods that can automatically determine the optimal set of diffusion steps to use for each dataset, thereby enhancing the performance of our directional diffusion models. This could involve techniques such as the adaptive selection of diffusion steps based on dataset characteristics.  Additionally, exploring the application of our method to computer vision and natural language processing tasks holds great potential for advancing  these domains. By adapting and extending our directional diffusion models to these areas, we may leverage their inherent strengths to improve representations and enable effective learning tasks such as image recognition, object detection, sentiment analysis, and language understanding.

\bibliographystyle{apalike}
\bibliography{cite}

\end{document}